\documentclass{article}

\usepackage{times}
\usepackage{graphicx} % more modern
\usepackage{subfig} 

\usepackage{natbib}

\usepackage{algorithm}
\usepackage{algorithmic}
\usepackage{amsmath}
\usepackage{url}
\usepackage{amsmath}
\usepackage{amssymb}
\usepackage{mathrsfs}
\usepackage{array}
\usepackage{bm}
\usepackage{multirow}
\usepackage{hyperref}
\usepackage{afterpage}

\usepackage{hyperref}

\usepackage[accepted]{arXivicml2016}

\def\E{{\rm E}}

\def\I{{\bf I}}
\def\w{{\bf w}}
\def\B{{\bf B}}
\def\b{{\bf b}}
\def\a{\alpha}
\def\tI{\tilde{\bf I}}
\def\tM{\tilde{M}}

\def\bF{{\bar{F}}}
\def\F{{\bf F}}

\def\D{{\cal D}}
\def\C{{\cal C}}
\def\S{{\cal S}}
\def\L{{\cal L}}

\def\obs{{\rm obs}}
\def\syn{{\rm syn}}

\newtheorem{Prop}{Proposition}
\newtheorem{Theo}{Theorem}
\newtheorem{Def}{Definition}

\icmltitlerunning{A Theory of Generative ConvNet}

\begin{document} 
%\afterpage{\cfoot{\thepage}}
\twocolumn[
\icmltitle{A Theory of Generative ConvNet}

% It is OKAY to include author information, even for blind
% submissions: the style file will automatically remove it for you
% unless you've provided the [accepted] option to the icml2016
% package.

\icmlauthor{Jianwen Xie $^\dagger$}{Jianwen@ucla.edu}
\icmlauthor{Yang Lu $^\dagger$}{yanglv@ucla.edu}
\icmlauthor{Song-Chun Zhu}{sczhu@stat.ucla.edu}
\icmlauthor{Ying Nian Wu}{ywu@stat.ucla.edu}
\icmladdress{Department of Statistics,
            University of California, Los Angeles, CA, USA}

% You may provide any keywords that you 
% find helpful for describing your paper; these are used to populate 
% the "keywords" metadata in the PDF but will not be shown in the document
\icmlkeywords{Auto-encoder, Generative models, Markov Random Field.}

\vskip 0.3in
]

\begin{abstract} 
We show that a generative random field model,  which we call generative ConvNet, can be derived from the commonly used discriminative ConvNet, by assuming a ConvNet for multi-category classification and assuming one of the categories is a base category generated by a reference distribution. If we further assume that the non-linearity in the ConvNet is Rectified Linear Unit (ReLU) and  the reference distribution is Gaussian white noise, then we obtain a generative ConvNet model that is unique among energy-based models: The model is piecewise Gaussian, and the means of the Gaussian pieces are defined by an auto-encoder, where the filters in the bottom-up encoding become the basis functions in the top-down decoding, and the binary activation variables detected by the filters in the bottom-up convolution process become the coefficients of the basis functions in the top-down deconvolution process. The Langevin dynamics for sampling the generative ConvNet is driven by the reconstruction error of this auto-encoder. The contrastive divergence learning of the generative ConvNet reconstructs the training images by the auto-encoder. The maximum likelihood learning algorithm can synthesize realistic natural image patterns. 
\end{abstract} 
\thispagestyle{firststyle}
\let\thefootnote\relax\footnotetext{$^\dagger$ Equal contributions.}

\section{Introduction}
\label{introduction}

The  convolutional neural network (ConvNet or CNN)  \citep{lecun1998gradient, krizhevsky2012imagenet} has proven to be a tremendously successful discriminative or predictive learning machine. Can  the discriminative ConvNet be turned into a generative model and an unsupervised learning machine? It would be highly desirable if this can be achieved because generative models and unsupervised learning can be very useful when the training datasets are small or the labeled data are scarce. It would also be extremely satisfying from a conceptual point of view if both discriminative classifier and generative model, and both supervised learning and unsupervised learning, can be treated within a unified framework for ConvNet. 

In this conceptual paper, we show that a generative random field model,  which we call generative ConvNet, can be derived from the commonly used discriminative ConvNet, by assuming a ConvNet for multi-category classification and assuming one of the categories is a base category generated by a reference distribution. The model is in the form of exponential tilting of the reference distribution, where the exponential tilting is defined by the ConvNet scoring function.  If we further assume that the non-linearity in the ConvNet is Rectified Linear Unit (ReLU)   \citep{krizhevsky2012imagenet} and that the reference distribution is Gaussian white noise, then we obtain a generative ConvNet model that is unique among energy-based models: The model is piecewise Gaussian, and the means of the Gaussian pieces are defined by an auto-encoder, where the filters in the bottom-up encoding become the basis functions in the top-down decoding, and the binary activation variables detected by the filters in the bottom-up convolution process become the coefficients of the basis functions in the top-down deconvolution process \cite{zeiler2013visualizing}.  The Langevin dynamics for sampling the generative ConvNet is driven by the reconstruction error of this auto-encoder. The contrastive divergence learning   \citep{Hinton2002a}  of the generative ConvNet reconstructs the training images by the auto-encoder. The maximum likelihood learning algorithm can synthesize realistic natural image patterns. 

The main purpose of our paper is to explore the properties of the generative ConvNet, such as being piecewise Gaussian with auto-encoding means, where the filters in the bottom-up operation take up a new role as the basis functions in the top-down representation. Such an internal representational structure is essentially unique among energy-based models.  They are the results of the marriage between the piecewise linear structure of the ReLU ConvNet and the Gaussian white noise reference distribution.  The auto-encoder we elucidate is a harmonious fusion of bottom-up convolution and top-down deconvolution. 

The reason we choose Gaussian white noise as the reference distribution is that it is the maximum entropy distribution with given marginal variance. Thus it is the most featureless distribution. Another justification for the Gaussian white noise distribution is that it is the limiting distribution if we zoom out any stochastic texture pattern,  due to the central limit theorem \cite{wu2004information}.  The ConvNet seeks to search for non-Gaussian features in order to recover the non-Gaussian distribution before the central limit theorem takes effect.  Without the Gaussian reference distribution that contributes the $\ell_2$ norm term to the energy function,  we will not have the auto-encoding form in the internal representation of the model. In fact, without Gaussian reference distribution, the density of the model is not even integrable. The Gaussian reference distribution is also crucial for the Langevin dynamics to be driven by the reconstruction error. 

The ReLU is the most commonly used non-linearity in modern ConvNet  \citep{krizhevsky2012imagenet}. It makes the ConvNet scoring function piecewise linear \cite{montufar2014number}.  Without the piecewise linearity of the ConvNet, we will not have the piecewise Gaussian form of the model. The ReLU is the source of the binary activation variables because the ReLU function can be written as $\max(0, r) = 1(r>0)*r$, where $1(r>0)$ is the binary variable indicating whether $r>0$ or not.  The binary variable can also be derived from the derivative of the ReLU function.  The piecewise linearity is also crucial for the exact equivalence between the gradient of the contrastive divergence  learning and the gradient of the auto-encoder reconstruction error.
% because it makes the tedious curvature term in score matching estimator  \cite{Hyvrinen05estimationof} disappear. 

\section{Related Work}

The model in the form of exponential tilting of a reference distribution  where the exponential tilting is defined by ConvNet was first proposed by \cite{Dai2015ICLR}. They did not study the internal representational structure of the model.  \cite{LuZhuWu2016} proposed to learn the FRAME (Filters, Random field, And Maximum Entropy) models  \citep{zhu1997minimax} based on the pre-learned filters of existing ConvNets. They did not learn the models from scratch. The hierarchical energy-based models \citep{Lecun2006} were studied by the pioneering work of \cite{hinton2006unsupervised} and \cite{Ng2011}. Their models do not correspond directly to modern ConvNet and do not posses the internal representational structure of the generative ConvNet. 

The generative ConvNet model can be  viewed as a hierarchical version of the FRAME model \citep{zhu1997minimax}, as well as the Product of Experts  \citep{Hinton2002a, teh2003energy} and  Field of Experts   \citep{roth2005fields} models. These models do not have explicit Gaussian white noise reference distribution. Thus they do not have the internal auto-encoding representation, and the filters in these models do not play the role of basis functions. In fact, a main motivation for this paper is to reconcile the FRAME model \citep{zhu1997minimax}, where the Gabor wavelets play the  role of bottom-up filters, and the Olshausen-Field model \citep{olshausen1997sparse}, where the wavelets play the  role of top-down basis functions.  The generative ConvNet may be seen as one step towards achieving this goal.  See also   \cite{xie2014learning}.

The relationship between energy-based model with latent variables and auto-encoder was discovered by  \cite{vincent2011connection} and \cite{Swersky2011} via the score matching estimator   \cite{Hyvrinen05estimationof}.  This connection requires that the free energy can be calculated analytically, i.e.,  the latent variables can be integrated out analytically. This is in general not the case for deep energy-based models with multiple layers of latent variables, such as deep Boltzmann machine with two layers of hidden units  \cite{salakhutdinov2009deep}. In this case, one cannot obtain an explicit auto-encoder. In fact, for such models, the inference of the latent variables is in general intractable. In generative ConvNet, the multiple layers of binary activation variables come from the ReLU units, and the means of the Gaussian pieces are always defined by an explicit hierarchical auto-encoder. 
 
Compared to hierarchical models with explicit binary latent variables such as those based on the Boltzmann machine \citep{Hinton06, salakhutdinov2009deep, lee2009convolutional}, the generative ConvNet is directly derived from the discriminative ConvNet. Our work seems to suggest that in searching for generative models and unsupervised learning machines, we need to look no further beyond the ConvNet. 

\section{Generative ConvNet} 

To fix notation, let $\I(x)$ be an image defined on the square (or rectangular) image domain $\D$, where $x = (x_1, x_2)$ indexes the coordinates of pixels. We can treat $\I(x)$ as a two-dimensional function defined on $\D$. We can also treat $\I$ as a vector if we fix an ordering for the pixels. 
For a filter $F$, let $F*\I$ denote the filtered image or feature map, and let $[F*\I](x)$ denote the filter response or feature at position $x$. 

A ConvNet is a composition of multiple layers of linear filtering and element-wise non-linear transformation as expressed by the following recursive formula: 
\begin{equation}
%\begin{aligned}
  [F^{(l)}_{k}  *\I](x)  =   h\left(\sum_{i=1}^{N_{l-1}}  \sum_{y \in \S_{l}} w^{(l, k)}_{i, y}   [F^{(l-1)}_{i}*\I](x+y) + b_{l, k}\right) \\
%\end{aligned}
\label{eq:ConvNet}
\end{equation}
where $l \in \{1, 2, ..., {\cal L}\}$ indexes the layer.  $\{F^{(l)}_k, k = 1, ..., N_l\}$ are the filters at layer $l$, and $\{F^{(l-1)}_i, i = 1, ..., N_{l-1}\}$ are the filters at layer $l-1$.  $k$ and $i$ are used to index filters at layers $l$ and $l-1$ respectively, and $N_l$ and $N_{l-1}$ are the numbers of filters at layers $l$ and $l-1$ respectively. The filters are locally supported, so the range of $y$  is within a local support $\S_{l}$  (such as a $7 \times 7$ image patch). At the bottom layer, $[F^{(0)}_k*\I](x) = \I_k(x)$, where $k \in \{R, G, B\}$ indexes the three color channels. Sub-sampling may be implemented so that in  $[F^{(l)}_{k}  *\I](x)$, $x \in \D_l \subset \D$.  For notational simplicity, we do not make local max pooling explicit in (\ref{eq:ConvNet}). 

We take $h(r) = \max(r, 0)$, the Rectified Linear Unit (ReLU), that is commonly adopted in modern ConvNet   \citep{krizhevsky2012imagenet}. 
%This crisp piecewise linear transformation is the root of the binary activation variables and piecewise Gaussian form of the model. But some results in this paper can be extended to general non-linearity. 

Let $(F^{(\L)}_k)$ be the top layer filters. The filtered images are usually $1 \times 1$ due to repeated sub-sampling. Suppose there are $C$ categories. For category $c \in \{1, ..., C\}$, the scoring function for classification is 
\begin{equation}
f_c(\I; w) = \sum_{k=1}^{N_\L} w_{c, k} [F^{(\L)}_k*\I], \label{eq:score0}
\end{equation}
 where $w_{c, k}$ are the category-specific weight parameters for classification. 
\begin{Def} 
Discriminative ConvNet: We define the following conditional distribution as the discriminative ConvNet: 
\begin{equation} 
   p(c|\I; w) = \frac{\exp[f_c(\I; w) + b_c]}{\sum_{c=1}^{C} \exp\left[f_c(\I; w)+b_c\right]}.  \label{eq:d}
\end{equation} 
where $b_c$ is the bias term, and $w$ collects all the weight and bias parameters at all the layers. 
\end{Def}
The discriminative ConvNet is a multinomial logistic regression (or soft-max) that is commonly used for classification  \citep{lecun1998gradient, krizhevsky2012imagenet}.

\begin{Def} 
Generative ConvNet (fully connected version): We define the following random field model as the fully connected version of  generative ConvNet: 
\begin{equation} 
  p(\I|c; w) =  p_c(\I; w) = \frac{1}{Z_c(w)} \exp[ f_c(\I; w)] q(\I), \label{eq:g}
\end{equation}
where $q(\I)$ is a reference distribution or the null model, assumed to be Gaussian white noise in this paper. $Z(w) = \E_q\{ \exp[ f_c(\I; w)] \}$ is the normalizing constant. 
\end{Def}
In (\ref{eq:g}), $p_c(\I; w)$ is obtained by the exponential tilting of $q(\I)$, and is the conditional distribution of image given category, $p(\I|c, w)$. The model was first proposed by \cite{Dai2015ICLR}.

\begin{Prop} \label{prop:1}
 Generative and discriminative ConvNets can be derived from each other: 
 
(a)  Let $\rho_c$ be the prior probability of category $c$, if $p(\I|c; w) =  p_c(\I; w)$ is defined according to model (\ref{eq:g}), then $p(c|\I; w)$  is given by model (\ref{eq:d}),  with $b_c = \log \rho_c - \log Z_c(w) + {\rm constant}$.

 (b) Suppose a base category $c = 1$ is generated by $q(\I)$, and suppose we fix  the scoring function and the bias term of the base category $f_1(\I; w) = 0$, and $ b_1 = 0$. If $p(c|\I; w)$ is given by model (\ref{eq:d}), then $p(\I|c; w) =  p_c(\I; w)$ is of the form of model $(\ref{eq:g})$, 
with $b_c = \log \rho_c - \log \rho_1 + \log Z_c(w)$.
\end{Prop}

%Proposition \ref{prop:1} can be proved by a simple exercise of the Bayes rule. Result (a) has already been explained in \cite{Dai2015ICLR}. Result (b) is stronger and is new. First, it is entirely reasonable to include Gaussian white noise images as a base category and demand the discriminative ConvNet (\ref{eq:d}) not to misclassify the Gaussian white noise as an object category. It is also reasonable to fix the scoring function and bias term of this base category at 0 in training for the sake of identifiability. In fact, in the binary (two-category) logistic regression, the scoring function and the bias term for the negative category are always fixed at 0.  Then 
%\begin{equation}
%\frac{p(c|\I; w)}{p(c=1|\I; w)} = \exp[f_c(\I; w) + b_c].
%\end{equation}
% Meanwhile 
% \begin{equation}
% \begin{aligned}
% \frac{p(c|\I; w)}{p(c=1|\I; w)}  = \frac{p(c, \I|w)}{p(c=1, \I|w)} = \frac{\rho_c p_c(\I; w)}{\rho_1 q(\I)}, 
% \end{aligned}
% \end{equation}
% because $p(c, \I|w) = p(c|\I; w) P(\I; w)$, where the marginal distribution $P(\I; w) = \sum_{c=1}^{C} \rho_c p(\I|c; w)$ is the mixture of all the categories.  Thus $p_c(\I; w)$ is of the form of model (\ref{eq:g}). 

%As to learning, we may use the discriminative log-likelihood based on $\log p(c|\I; w)$, or we may use the generative log-likelihood based on $\log p(c, \I | w) = \log p_c(\I; w) + \log \rho_c$. Because $\log p(c, \I | w) = \log p(c|\I; w) + \log P(\I; w)$, the discriminative log-likelihood $\log p(c|\I; w)$  is without the marginal log-likelihood $\log P(\I; w)$,   resulting in the loss of statistical efficiency. 

If we only observe unlabeled data $\{\I_m, m = 1, ..., M\}$, we may still use the exponential tilting form to model and learn from them. A possible model is to 
learn filters at a certain convolutional layer $L \in \{1, ..., \L\}$ of a ConvNet. 
\begin{Def} Generative ConvNet (convolutional version): we define the following Markov random field model as the convolutional version of generative ConvNet: 
\begin{equation}
p(\I; w) = \frac{1}{Z(w)} \exp \left[\sum_{k=1}^{K} \sum_{x \in {\cal D}_L} [F_k^{(L)}*\I](x)\right] q(\I), 
\label{eq:ConvNet-FRAME}
\end{equation}
where $w$ consists of all the weight and bias terms that define the filters $(F_k^{(L)}, k = 1, ..., K = N_L)$, and $q(\I)$ is the Gaussian white noise model. 
\end{Def}

Model (\ref{eq:ConvNet-FRAME})  corresponds to the exponential tilting model (\ref{eq:g}) with scoring function
\begin{equation}
f(\I; w) = \sum_{k=1}^{K} \sum_{x \in {\cal D}_L} [F_k^{(L)}*\I](x). \label{eq:a}
\end{equation}
%Essentially the above model treats the images $\{\I_m\}$ as coming from a single meta-category, which is to be discriminated from the base category $q$ by the filters $(F^{(L)}_k)$ to be learned from the data. However, it may be too easy to discriminate $\{\I_m\}$ from $q$, so that we cannot learn anything meaningful by the discriminative log-likelihood.  In this case, we can learn $w$ based on the generative log-likelihood  $L(w) = \sum_{m=1}^{M} \log p(\I_m; w)/M$, with $p(\I; w)$ defined by (\ref{eq:ConvNet-FRAME}). The learning of filters $\{F_k^{(L)}\}$ by the generative log-likelihood is considered to be unsupervised because the observed images are unlabeled. 
For the rest of the paper, we shall focus on the model (\ref{eq:ConvNet-FRAME}),  but all the results can be easily extended to model (\ref{eq:g}).

\section{A Prototype Model} 

In order to reveal the internal structure of  the generative ConvNet, it helps to start from the simplest prototype model. A similar model was studied by  \cite{xie2015boosting}. 
%The generative ConvNet can be obtained from the prototype model by unfolding the latter both convolutionally and hierarchically
%, but with much more involved notation that is in danger of obscuring the key ideas. Hence it is helpful to start from the prototype model. 

In our prototype model, we assume that the image domain $\D$ is small (e.g., $10 \times 10$). Suppose we want to learn a dictionary of filters or basis functions from a set of observed image patches $\{\I_m, m = 1, ..., M\}$ defined on $\D$. We denote these filters or basis functions by $(\w_k, k = 1, ..., K)$, where each $\w_k$ itself is an image patch defined on $\D$. Let $\langle \I, \w_k\rangle = \sum_{x\in \D} \w_k(x) \I(x)$ be the inner product between image patches $\I$ and $\w_k$.  It is also the response of $\I$ to the linear filter $\w_k$. 

\begin{Def} Prototype model: We define the following random field model as the prototype model:
\begin{equation} 
  p(\I; w) = \frac{1}{Z(w)} \exp\left[ \sum_{k=1}^{K} h(\langle \I, \w_k\rangle + b_k)\right] q(\I), \label{eq:p}
\end{equation} 
where $b_k$ is the bias term, $w = (\w_k, b_k, k = 1, ..., K)$, and $h(r) = \max(r, 0)$. $q(\I)$ is the Gaussian white noise model, 
\begin{equation}
q(\I) = \frac{1}{(2\pi\sigma^2)^{|{\cal D}|/2}} \exp\left[- \frac{1}{2\sigma^2} ||\I||^2\right], 
\label{eq:Gaussian}
\end{equation}   
where $|\D|$ counts the number of pixels in the domain $\D$. 
\end{Def}

%The following are the properties of the prototype model. 

{\em Piecewise Gaussian and binary activation variables:}  Without loss of generality, let us assume $\sigma^2 = 1$ in $q(\I)$. Define the binary activation variable $\delta_k(\I; w) = 1$ if $\langle \I, \w_k\rangle + b_k>0$ and $\delta_k(\I; w) = 0$ otherwise, i.e., 
%\begin{equation}
$\delta_k(\I; w) = 1(\langle \I, \w_k\rangle + b_k > 0)$,
%\end{equation}
 where $1()$ is the indicator function. Then 
%\begin{equation}
$h(\langle\I, \w_k\rangle + b_k) = \delta_k(\I; w) (\langle\I, \w_k\rangle+b_k)$.
%\end{equation}
  The image space is divided into $2^{K}$ pieces by the $K$ hyper-planes, $\langle \I, \w_k\rangle + b_k  = 0$, $k = 1, ..., K$, according to the values of the binary activation variables $(\delta_k(\I; w), k = 1, ..., K)$. Consider the piece of image space where $\delta_k(\I; w) = \delta_k$ for $k = 1, ..., K$. Here we abuse the notation slightly where $\delta_k \in \{0, 1\}$ on the right hand side denotes the value of $\delta_k(\I; w)$. Write $\delta(\I; w) = (\delta_k(\I; w), k = 1, ..., K)$, and $\delta = (\delta_k, k = 1, ..., K)$ as an instantiation of $\delta(\I; w)$. We call $\delta(\I; w)$ the activation pattern of $\I$. Let $A(\delta; w) = \{\I: \delta(\I; w) = \delta\}$ be the piece of image space that consists of images sharing the same activation pattern $\delta$, then the probability density on this piece
\begin{equation} 
\begin{aligned}
   p(\I; w, \delta) &\propto \exp\left[ \sum_{k=1}^{K} \delta_k b_k +  \langle \I, \sum_{k=1}^{K} \delta_k \w_k\rangle - \frac{\|\I\|^2}{2}\right] \\
     & \propto \exp \left[-\frac{1}{2} \|\I - \sum_{k=1}^{K} \delta_k \w_k\|^2\right], \label{eq:n}
  \end{aligned}
\end{equation}
which  is ${\rm N}(\sum_k \delta_k \w_k, {\bf 1})$ restricted to the piece $A(\delta; w)$, where the bold font ${\bf 1}$ is the identity matrix.
% (recall we assume $\sigma^2 = 1$). 
% $\delta = (\delta_k)$ are the binary activation variables that generate
  The mean of this Gaussian piece
%  , $\sum_k \delta_k \w_k$,   
  seeks to reconstruct images in $A(\delta; w)$ via the auto-encoding scheme  $\I \rightarrow \delta \rightarrow \sum_k \delta_k \w_k$. 

 {\em Synthesis via reconstruction:} One can sample from $p(\I; w)$ in (\ref{eq:p}) by the Langevin dynamics: 
\begin{equation}
\I_{\tau+1} = \I_{\tau} - \frac{\epsilon^2}{2} \left[\I_\tau - \sum_{k=1}^{K}\delta_k(\I_\tau; w) \w_k\right] + \epsilon Z_\tau,
\end{equation}
 where $\tau$ denotes the time step, $\epsilon$ denotes the step size, assumed to be sufficiently small throughout this paper, and $Z_\tau \sim {\rm N}(0, {\bf 1})$. The dynamics is driven by the auto-encoding reconstruction error $\I_\tau - \sum_{k=1}^{K}\delta_k(\I_\tau; w) \w_k$, where the reconstruction is based on the binary activation variables $(\delta_k)$. 
This links synthesis to reconstruction. 
 
{\em Local modes are exactly auto-encoding:} If the mean of a Gaussian piece is a local energy minimum $\hat{\I}$, we have the exact auto-encoding
%\begin{equation}
$\hat{\I} = \sum_{k=1}^{K} \delta_k(\hat{\I}; w) \w_k$. 
%\end{equation}
The encoding process is bottom-up and infers $\delta_k = \delta_k(\hat{\I}; w) = 1(\langle \hat{\I}, \w_k\rangle + b_k > 0)$.  The decoding process is top-down and reconstructs $\hat{\I} = \sum_k \delta_k \w_k$. In the encoding process, $\w_k$ plays the role of filter. In the decoding process, $\w_k$ plays the role of basis function. 
  
\section{Internal Structure of Generative ConvNet} 

In order to generalize the prototype model (\ref{eq:p}) to the generative ConvNet (\ref{eq:ConvNet-FRAME}), we only need to add two elements: (1) Horizontal unfolding: make the filters $(\w_k)$ convolutional. (2) Vertical unfolding: make the filters $(\w_k)$ multi-layer or hierarchical. 
The results we have obtained for the prototype model can be unfolded accordingly.  

To derive the internal representational structure of the generative ConvNet, the key is to write the scoring function  $f(\I; w)$ as a linear function $\alpha + \langle \I, \B\rangle$ on each piece of image space with fixed activation pattern. Combined with the $\|\I\|^2/2$ term from $q(\I)$, the energy function will be quadratic, i.e., $\|\I\|^2/2 - \langle \I, \B\rangle - \alpha$, and the probability distribution will be truncated Gaussian with $\B$ being the mean. In order to write $f(\I; w) = \alpha + \langle \I, \B\rangle$ for fixed activation pattern, we  shall use vector notation for ConvNet, and derive $\B$ by a top-down deconvolution process.  $\B$ can also be obtained by $\partial f(\I; w)/\partial \I$ via back-propagation computation. 

For filters at level $l$, the $N_l$ filters are denoted by the compact notation $\F^{(l)} = (F_k^{(l)}, k = 1, ..., N_l)$. The $N_l$ filtered images or feature maps are denoted by the compact notation $\F^{(l)}*\I = (F_k^{(l)}*\I, k = 1, ..., N_l)$. $\F^{(l)}*\I$ is a 3D image, containing all the $N_l$ filtered images at layer $l$. In vector notation,  the recursive formula (\ref{eq:ConvNet}) of ConvNet filters can be rewritten as 
\begin{equation}
 [F^{(l)}_{k}*\I](x) = h\left(\langle \w^{(l)}_{k, x}, \F^{(l-1)}*\I \rangle + b_{l, k}\right), \label{eq:v}
\end{equation}
where $\w^{(l)}_{k, x}$ matches the dimension of $\F^{(l-1)}*\I$, which is a 3D image containing all the $N_{l-1}$ filtered images at layer $l-1$. Specifically, 
\begin{equation}
\langle \w^{(l)}_{k, x}, \F^{(l-1)}*\I \rangle = \sum_{i=1}^{N_{l-1}}  \sum_{y \in \S_{l}} w^{(l, k)}_{i, y}   [F^{(l-1)}_{i}*\I](x+y).
\end{equation} 
The 3D basis functions $(\w^{(l)}_{k, x})$ are locally supported, and they are spatially translated copies for different positions $x$, i.e., 
%\begin{equation}
$\w^{(l)}_{k, x, i}(x+y) = w^{(l, k)}_{i, y}$,
%\end{equation}
 for $i \in \{1, ..., N_{l-1}\}$,  $x \in \D_{l}$ and $y \in \S_l$. For instance, at layer $l = 1$, $\w^{(1)}_{k, x}$ is a Gabor-like wavelet of type $k$ centered at position $x$. 

%$\w^{(l)}_{k, x}$ is the unfolded version of $\w_k$ in the prototype model, where $x$ indexes the position for convolutional unfolding, and $l$ indexes the layer for hierarchical unfolding. 

Define the binary activation variable
\begin{equation}
\delta_{k, x}^{(l)}(\I; w) = 1\left(\langle \w^{(l)}_{k, x}, \F^{(l-1)}*\I \rangle + b_{l, k} > 0\right). \label{eq:bua}
\end{equation}
% Since $F^{(l)}_k$ corresponds to a non-stationary FRAME model (\ref{eq:iFRAME}), $\delta_{k, x}^{(l)}(\I; w)$ is a decision maker based on the likelihood ratio test of $H_1: p_k^{(l)}(\I; w, x)$ vs $H_0: q(\I)$ for detecting the pattern modeled by $F^{(l)}_k$. 

According to  (\ref{eq:ConvNet}), we have the following bottom-up process: 
\begin{equation}
%\begin{aligned}
%\mbox{\em Bottom-up}:  
[F^{(l)}_{k}  *\I](x)   =\delta_{k, x}^{(l)}(\I; w)  \left(\langle \w^{(l)}_{k, x}, \F^{(l-1)}*\I \rangle + b_{l, k}\right). 
%\end{aligned}
\label{eq:ConvNet1}
\end{equation}

Let $\delta(\I; w) = (\delta^{(l)}_{k, x}(\I; w), \forall k, x, l)$ be the activation pattern at all the layers. 
The active pattern $\delta(\I; w)$ can be computed in the bottom-up process (\ref{eq:bua}) and (\ref{eq:ConvNet1}) of ConvNet.   

For  the scoring function  $f(\I; w) = \sum_{k=1}^{K} \sum_{x \in {\cal D}_L} [F_k^{(L)}*\I](x)$  defined  in (\ref{eq:a}) for the generative ConvNet, we can write it in terms of lower layers $(l \leq L)$ of filter responses: 
\begin{equation}
\begin{aligned}
   f(\I; w) & = \alpha_l + \langle \B^{(l)}, \F^{(l)}*\I\rangle\\
     &= \alpha_l + \sum_{k=1}^{N_l} \sum_{x \in \D_l} \B^{(l)}_k(x) [F^{(l)}_k*\I](x),  \label{eq:f}
\end{aligned}
\end{equation}
where $\B^{(l)} = (\B^{(l)}_k(x), k = 1, ..., N_l, x \in \D_l)$ consists of the maps of the coefficients at layer $l$. $\B^{(l)}$  matches the dimension of $\F^{(l)}*\I$. When $l=L$, $\B^{(L)}$ consists of maps of 1's, i.e., $\B^{(L)}_k(x) = 1$ for $k = 1, ..., K = N_L$ and $x \in \D_L$. 
According to equations (\ref{eq:ConvNet1}) and (\ref{eq:f}),  we have the following top-down process: 
\begin{equation}
%\mbox{\em Top-down}:    
\B^{(l-1)} =   \sum_{k=1}^{N_l} \sum_{x\in \D_l}  \B^{(l)}_k(x)   \delta^{(l)}_{k, x} (\I; w) \w^{(l)}_{k, x}, \label{eq:de}
\end{equation}
where both $\B^{(l-1)}$ and $ \w^{(l)}_{k, x}$ match the dimension of $\F^{(l-1)}*\I$. 
Equation (\ref{eq:de}) is a top-down deconvolution process, where $\B^{(l)}_k(x)   \delta^{(l)}_{k, x}$ serves as the coefficient of  the basis function $\w^{(l)}_{k, x}$. The top-down deconvolution process (\ref{eq:de}) is similar to but subtly different from that in \cite{zeiler2013visualizing}, because equation (\ref{eq:de}) is controlled by the multiple layers of activation variables   $ \delta^{(l)}_{k, x}$  computed in the bottom-up process of the ConvNet. Specifically, $\delta^{(l)}_{k, x}$ turns on or off the basis function $\w^{(l)}_{k, x}$, while $\delta^{(l)}_{k, x}$ is determined by $F^{(l)}_{k}$. The recursive relationship for $\alpha_l$ can be similarly derived. 

In the bottom-up convolution process (\ref{eq:ConvNet1}), $(\w^{(l)}_{k, x})$ serve as filters. In the top-down deconvolution process (\ref{eq:de}), $(\w^{(l)}_{k, x})$ serve as basis functions. 
%The top-down deconvolution process (\ref{eq:de}) defines a recursive reconfigurable compositional structure, where the reconfiguration is controlled by the binary activation variables. 

Let $\B = \B^{(0)}$, $\alpha = \alpha_0$. Since $\F^{(0)}*\I = \I$, we have $f(\I; w) = \alpha + \langle \I, \B\rangle$. Note that $\B$ depends on the activation pattern $\delta(\I; w) = (\delta^{(l)}_{k, x}(\I; w), \forall k, x, l)$,  as well as $w$ that collects the weight and bias parameters at all the layers. 

On the piece of image space $A(\delta; w) = \{\I: \delta(\I; w) = \delta\}$ of a fixed activation pattern (again we slightly abuse the notation where $\delta = (\delta_{k, x}^{(l)} \in \{0, 1\}, \forall k, x, l)$ denotes an instantiation of the activation pattern), $\B$ and $\alpha$ depend on $\delta$ and $w$. To make this dependency explicit, we denote $\B = \B_{w, \delta}$ and $\alpha = \alpha_{w, \delta}$, thus
\begin{equation}
    f(\I; w) = \a_{w, \delta}+ \langle \I,  \B_{w, \delta}\rangle.
\end{equation}  
See \cite{montufar2014number} for an analysis of the number of linear pieces. 

\begin{Theo} \label{theo:1}
Generative ConvNet is piecewise Gaussian: With ReLU $h(r) = \max(0, r)$ and Gaussian white noise $q(\I)$, $p(\I; w)$ of model (\ref{eq:ConvNet-FRAME}) is piecewise Gaussian. On each piece $A(\delta; w)$, the density is  ${\rm N}(\B_{w, \delta}, {\bf 1})$ truncated to $A(\delta; w)$, i.e., $\B_{w, \delta}$ is an approximated reconstruction of images in $A(\delta; w)$. 
\end{Theo}

Theorem  \ref{theo:1} follows from the fact that on $A(\delta; w)$, 
\begin{equation} 
\begin{aligned}
   p(\I; w, \delta) &\propto \exp\left[ \a_{w, \delta} +  \langle \I, \B_{w, \delta}\rangle - \frac{\|\I\|^2}{2}\right] \\
     & \propto \exp \left[-\frac{1}{2} \|\I - \B_{w, \delta}\|^2\right], \label{eq:G}
  \end{aligned}
\end{equation}
which is ${\rm N}(\B_{w, \delta}, {\bf 1})$ restricted to $A(\delta; w)$. 

For each $\I$, the binary activation variables in  $\delta = \delta(\I; w)$ are computed by the bottom-up convolution process  (\ref{eq:bua}) and (\ref{eq:ConvNet1}), and $\B_{w, \delta}$ is computed by the top-down deconvolution process (\ref{eq:de}). $\B_{w, \delta}$ seeks to reconstruct images in $A(\delta; w)$ via the auto-encoding scheme $\I \rightarrow \delta \rightarrow \B_{w, \delta}$. 

One can sample from $p(\I; w)$  of model (\ref{eq:ConvNet-FRAME}) by the Langevin dynamics: 
\begin{equation}
\I_{\tau+1} = \I_{\tau} - \frac{\epsilon^2}{2} \left[\I_\tau - \B_{w, \delta(\I_\tau; w)}\right] + \epsilon Z_\tau, \label{eq:Langevin}
\end{equation}
 where $Z_\tau \sim {\rm N}(0, {\bf 1})$. Again,  the dynamics is driven by the auto-encoding reconstruction error $\I -\B_{w, \delta(\I; w)}$.   
 
The deterministic part of the Langevin equation (\ref{eq:Langevin}) defines an attractor dynamics that converges to a local energy minimum \citep{zhu1997GRADE}.
 \begin{Prop} \label{prop:3}
 The local modes are exactly auto-encoding: 
Let $\hat{\I}$ be a local maximum of $p(\I; w)$ of model (\ref{eq:ConvNet-FRAME}), then we have exact auto-encoding of $\hat{\I}$  with the following bottom-up and top-down passes: 
 \begin{equation} 
 \begin{aligned}
 \mbox{Bottom-up encoding:} \; & \delta = \delta(\hat{\I}; w);   \\
 \mbox{Top-down decoding:} \;& \hat{\I}  = \B_{w, \delta}. \label{eq:decoding}
 \end{aligned}
 \end{equation} 
\end{Prop}
The local energy minima are the means of the Gaussian pieces in Theorem  \ref{theo:1},  but the reverse is not necessarily true because $\B_{w, \delta}$ does not necessarily belong to $A(\delta; w)$. But if $\B_{w, \delta} \in A(\delta; w)$, then $\B_{w, \delta}$ must be a local mode and is exactly auto-encoding. 

Proposition \ref{prop:3} can be generalized to general non-linear $h()$, whereas Theorem  \ref{theo:1}  is true only for piecewise linear $h()$ such as ReLU. 

Proposition \ref{prop:3} is  related to the Hopfield network \citep{hopfield1982neural}. 
%and attractor network  \citep{Seung98learningcontinuous}, 
It shows that the the Hopfield minima can be represented by a hierarchical auto-encoder.

\section{Learning Generative ConvNet}

The learning of $w$ from training images $\{\I_m, m = 1, ..., M\}$ can be accomplished by maximum likelihood. Let 
%\begin{equation}
$L(w) = \sum_{m=1}^{M} \log p(\I; w)/M$,
%\end{equation}
with $p(\I; w)$ defined in (\ref{eq:ConvNet-FRAME}), then
\begin{equation}
%\begin{aligned}
\frac{\partial L(w)}{\partial w} = \frac{1}{M} \sum_{m=1}^{M} \frac{\partial}{\partial w} f(\I_m; w) 
   -  \E_{w} \left[ \frac{\partial}{\partial w} f(\I; w) \right].
%\end{aligned}
\end{equation}

The expectation can be approximated by Monte Carlo samples \citep{younes1999convergence} from the Langevin dynamics (\ref{eq:Langevin}). See Algorithm \ref{code:FRAME}  for a description of the learning and sampling algorithm.

\begin{algorithm}
\caption{Learning and sampling algorithm}
\label{code:FRAME}
\begin{algorithmic}[1]

\REQUIRE ~~\\
(1)  training images $\{\I_m, m=1,...,M\}$ \\
(2) number of synthesized images $\tilde{M}$\\
(3) number of Langevin steps $L$\\
(4) number of learning iterations $T$

\ENSURE~~\\
(1) estimated parameters $w$\\
(2) synthesized images $\{\tI_m, m = 1, ..., \tilde{M}\}$ 

\item[]
\STATE Let $t\leftarrow 0$, initialize $w^{(0)} \leftarrow 0$.
\STATE Initialize $\tI_m \leftarrow 0$, for $m = 1, ..., \tilde{M}$. 
\REPEAT 
\STATE For each $m$, run $L$ steps of Langevin dynamics to update $\tI_m$, i.e., starting from the current $\tI_m$, each step 
follows equation (\ref{eq:Langevin}). 
\STATE Calculate  $H^{\obs} = \sum_{m=1}^{M} \frac{\partial}{\partial w} f(\I_m; w^{(t)})/M$, and
$H^{\syn} =  \sum_{m=1}^{\tM} \frac{\partial}{\partial w} f(\tI_m; w^{(t)})/\tM$.
\STATE Update $w^{(t+1)} \leftarrow w^{(t)} + \eta ( H^{\obs} - H^{\syn}) $,  with step size $\eta$. 
\STATE Let $t \leftarrow t+1$
\UNTIL $t = T$
\end{algorithmic}
\end{algorithm}

If we want to learn from big data,  we may use the contrastive divergence \citep{Hinton2002a} by starting the Langevin dynamics from the 
observed images.  The contrastive divergence tends to learn the auto-encoder in the generative ConvNet. 

Suppose we start from an observed image  $\I^{\obs}$, and run a small number of iterations of Langevin dynamics (\ref{eq:Langevin}) to get a synthesized image  $\I^{\syn}$. If 
both $\I^{\obs}$ and $\I^{\syn}$ share the same activation pattern $\delta(\I^{\obs}; w) = \delta(\I^{\syn}; w) = \delta$, then $f(\I; w) = a_{w, \delta} + \langle \I, \B_{w, \delta}\rangle$ for 
both $\I^{\obs}$ and $\I^{\syn}$. Then the contribution of $\I^{\obs}$ to the learning gradient is 
\begin{equation}
\frac{\partial}{\partial w} f(\I^{\obs}; w) - \frac{\partial}{\partial w} f(\I^{\syn}; w) =  \langle \I^{\obs} - \I^{\syn}, \frac{\partial}{\partial w} \B_{w, \delta}\rangle. \label{eq:c1}
\end{equation}
If $\I^{\syn}$ is close to the mean $\B_{w, \delta}$ and if  $\B_{w, \delta}$ is a local mode, then the contrastive divergence tends to reconstruct $\I^{\obs}$ by the local mode $\B_{w, \delta}$,   because the gradient 
\begin{equation}
 \frac{\partial}{\partial w}\|\I^{\obs} - \B_{w, \delta}\|^2/2 =-  \langle\I^{\obs} - \B_{w, \delta},  \frac{\partial}{\partial w} \B_{w, \delta}\rangle. \label{eq:c2}
\end{equation}
Hence contrastive divergence learns Hopfield network which memorizes observations by local modes.

\begin{figure}
	\centering
	\setlength{\fboxrule}{1pt}
	\setlength{\fboxsep}{0cm}	
	\subfloat{
		\includegraphics[width=.33\linewidth]{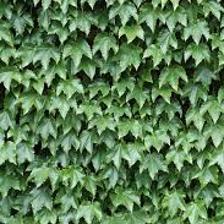}
		\includegraphics[width=.33\linewidth]{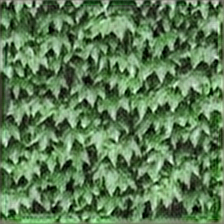}
		\includegraphics[width=.33\linewidth]{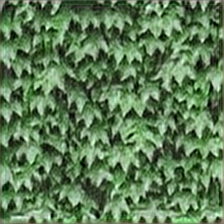}
	}\\[3px]
	\subfloat{
		\includegraphics[width=.33\linewidth]{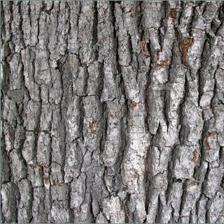}
		\includegraphics[width=.33\linewidth]{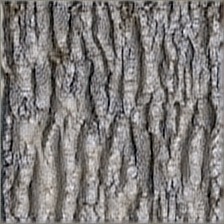}
		\includegraphics[width=.33\linewidth]{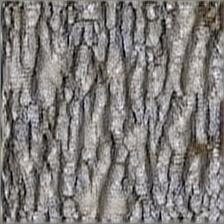}
	}\\[3px]
	\subfloat{
		\includegraphics[width=.33\linewidth]{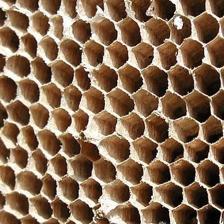}
		\includegraphics[width=.33\linewidth]{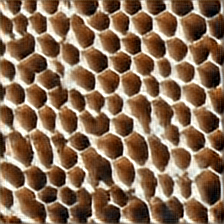}
		\includegraphics[width=.33\linewidth]{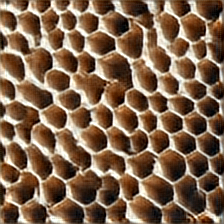}
	}\\[3px]
	\subfloat{
		\includegraphics[width=.33\linewidth]{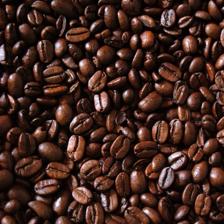}
		\includegraphics[width=.33\linewidth]{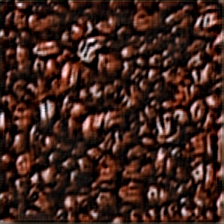}
		\includegraphics[width=.33\linewidth]{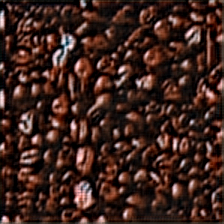}
	}\\[3px]
	\subfloat{
		\includegraphics[width=.33\linewidth]{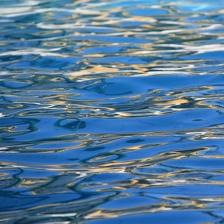}
		\includegraphics[width=.33\linewidth]{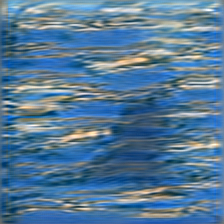}
		\includegraphics[width=.33\linewidth]{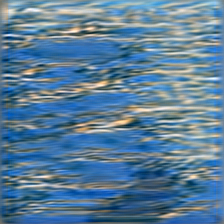}
	}\\[3px]
	\subfloat{
		\includegraphics[width=.33\linewidth]{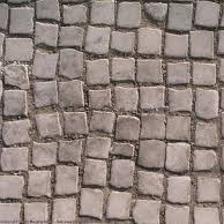}
		\includegraphics[width=.33\linewidth]{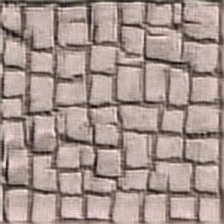}
		\includegraphics[width=.33\linewidth]{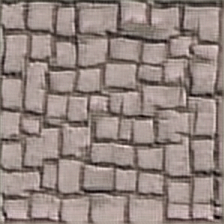}
	}\\[3px]		
		\subfloat{
		\includegraphics[width=.33\linewidth]{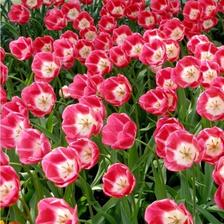}
		\includegraphics[width=.33\linewidth]{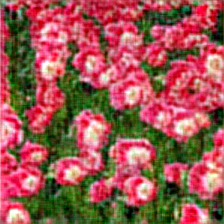}
		\includegraphics[width=.33\linewidth]{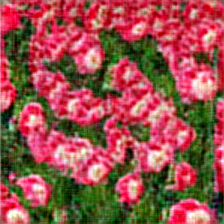}
	}
	\caption{Generating texture patterns. For each category, the first image is the training image, and the rest are 2 of the images generated by the learning algorithm.}
	\label{fig:texture}
\end{figure}

\begin{figure}
	\centering
	\setlength{\fboxrule}{1pt}
	\setlength{\fboxsep}{0cm}
	
		\subfloat{
		\includegraphics[width=.23\linewidth]{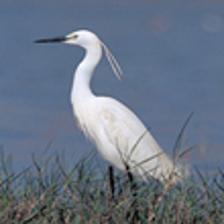}
		\includegraphics[width=.23\linewidth]{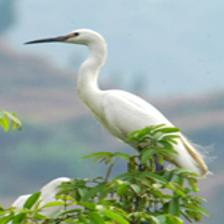}
		\includegraphics[width=.23\linewidth]{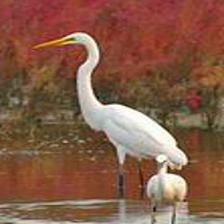}
		\includegraphics[width=.23\linewidth]{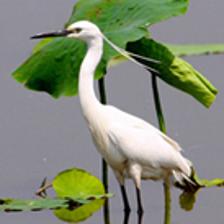}
	}\\[1px]
	\subfloat{
		\includegraphics[width=.23\linewidth]{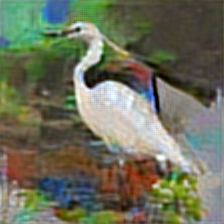}
		\includegraphics[width=.23\linewidth]{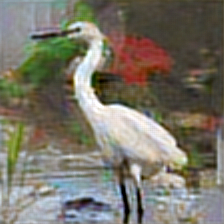}
		\includegraphics[width=.23\linewidth]{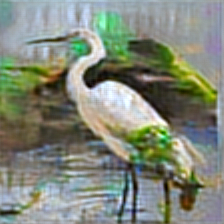}
		\includegraphics[width=.23\linewidth]{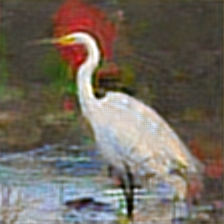}
	}\\[3px]	
	
	\subfloat{
		\includegraphics[width=.23\linewidth]{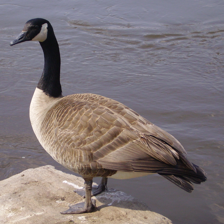}
		\includegraphics[width=.23\linewidth]{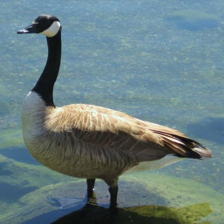}
		\includegraphics[width=.23\linewidth]{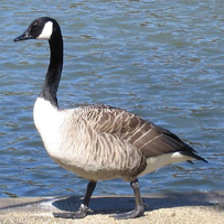}
		\includegraphics[width=.23\linewidth]{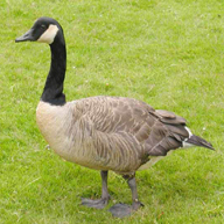}
	}\\[1px]
	\subfloat{
		\includegraphics[width=.23\linewidth]{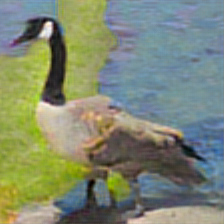}
		\includegraphics[width=.23\linewidth]{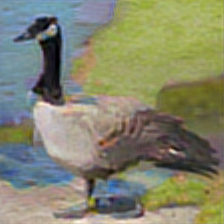}
		\includegraphics[width=.23\linewidth]{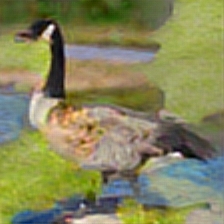}
		\includegraphics[width=.23\linewidth]{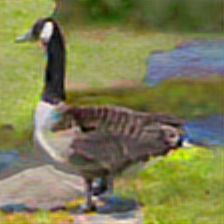}
	}\\[3px]
	
	\subfloat{
		\includegraphics[width=.23\linewidth]{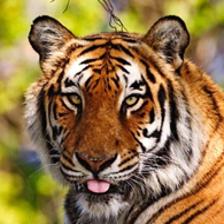}
		\includegraphics[width=.23\linewidth]{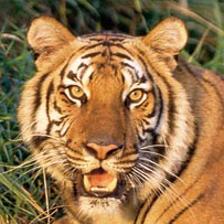}
		\includegraphics[width=.23\linewidth]{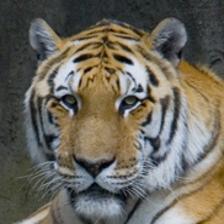}
		\includegraphics[width=.23\linewidth]{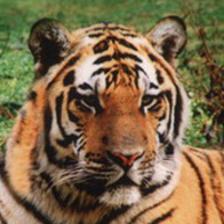}
	}\\[1px]
	\subfloat{
		\includegraphics[width=.23\linewidth]{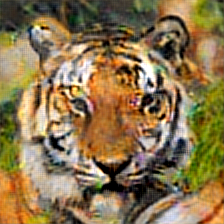}
		\includegraphics[width=.23\linewidth]{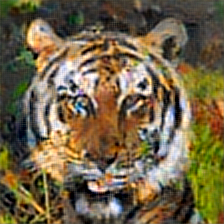}
		\includegraphics[width=.23\linewidth]{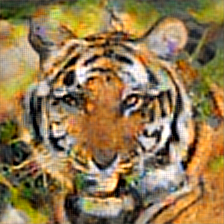}
		\includegraphics[width=.23\linewidth]{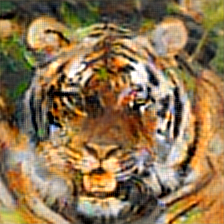}
	}\\[3px]

	\subfloat{
		\includegraphics[width=.23\linewidth]{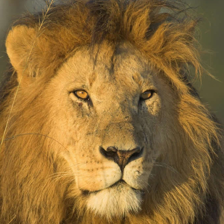}
		\includegraphics[width=.23\linewidth]{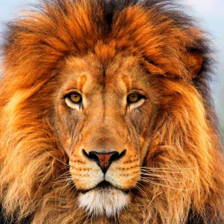}
		\includegraphics[width=.23\linewidth]{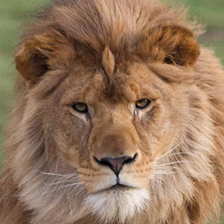}
		\includegraphics[width=.23\linewidth]{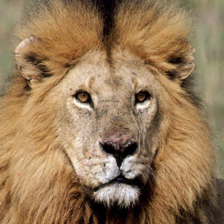}
	}\\[1px]
	\subfloat{
		\includegraphics[width=.23\linewidth]{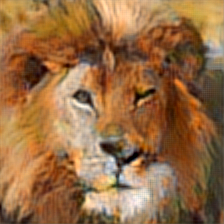}
		\includegraphics[width=.23\linewidth]{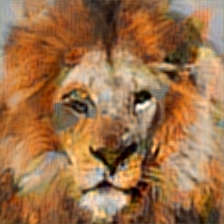}
		\includegraphics[width=.23\linewidth]{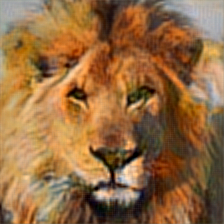}
		\includegraphics[width=.23\linewidth]{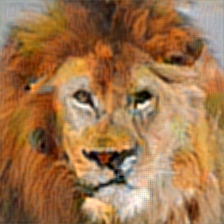}
	}\\[3px]
	
	%\subfloat{
	%	\includegraphics[width=.23\linewidth]{./figures/object/eight/train0001.png}
	%	\includegraphics[width=.23\linewidth]{./figures/object/eight/train0002.png}
	%	\includegraphics[width=.23\linewidth]{./figures/object/eight/train0003.png}
	%	\includegraphics[width=.23\linewidth]{./figures/object/eight/train0005.png}
	%}\\[1px]
	%\subfloat{
	%	\includegraphics[width=.23\linewidth]{./figures/object/eight/layer_03_001.png}
	%	\includegraphics[width=.23\linewidth]{./figures/object/eight/layer_03_004.png}
	%	\includegraphics[width=.23\linewidth]{./figures/object/eight/layer_03_005.png}
	%	\includegraphics[width=.23\linewidth]{./figures/object/eight/layer_03_013.png}
	%}
	
	\subfloat{
		\includegraphics[width=.23\linewidth]{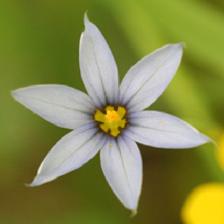}
		\includegraphics[width=.23\linewidth]{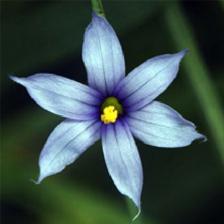}
		\includegraphics[width=.23\linewidth]{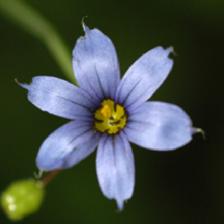}
		\includegraphics[width=.23\linewidth]{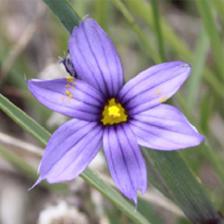}
	}\\[1px]
	\subfloat{
		\includegraphics[width=.23\linewidth]{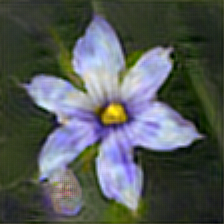}
		\includegraphics[width=.23\linewidth]{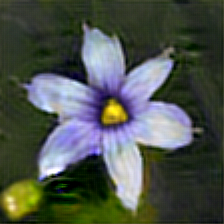}
		\includegraphics[width=.23\linewidth]{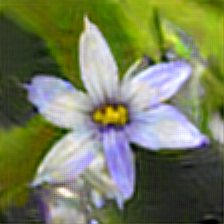}
		\includegraphics[width=.23\linewidth]{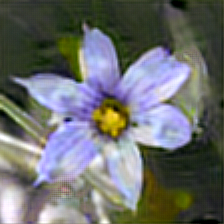}
	}			
	\caption{Generating object patterns. For each category, the first row displays 4 of the training images, and the second row displays 4 of the images generated by the learning algorithm.}
	\label{fig:object}
\end{figure}

 We can establish a precise connection for one-step contrastive divergence. 
 \begin{Prop} \label{prop:4}
Contrastive divergence learns to reconstruct: If the one-step Langevin dynamics does not change the activation pattern, i.e., $\delta(\I^{\obs}; w) = \delta(\I^{\syn}; w) = \delta$,  then the one-step contrastive divergence has an expected gradient that is proportional to the reconstruction gradient:
\begin{equation}
\E\left[ \frac{\partial}{\partial w} f(\I^{\obs}; w) - \frac{\partial}{\partial w} f(\I^{\syn}; w)\right] \propto \frac{\partial}{\partial w}\|\I^{\obs} - \B_{w, \delta}\|^2. 
\end{equation}
 \end{Prop}
This is because 
%\begin{equation}
$\I^{\syn} = \I^{\obs} - \frac{\epsilon^2}{2} \left[\I^{\obs} - \B_{w, \delta}\right] + \epsilon Z$,
%\end{equation}
 hence
% \begin{equation}
$ \E_Z\left[\I^{\obs} - \I^{\syn}\right]  \propto \I^{\obs} - \B_{w, \delta}$,
% \end{equation}
  and Proposition \ref{prop:4}  follows from (\ref{eq:c1}) and (\ref{eq:c2}). 

The contrastive divergence learning updates the bias terms to match the statistics of the activation patterns of $\I^{\obs}$ and $\I^{\syn}$, which helps to ensure that $\delta(\I^{\obs}; w) = \delta(\I^{\syn}; w)$. 

Proposition \ref{prop:4} is related to score matching estimator  \cite{Hyvrinen05estimationof}, whose connection with contrastive divergence based on one-step Langevin was studied by \cite{hyvarinen2007connections}. 
%The relationship between score matching and auto-encoder was discovered by  \cite{vincent2011connection} and \cite{Swersky2011}. 
Our work can be considered a sharpened specialization of this connection, where the piecewise linear structure in ConvNet greatly simplifies the matter by getting rid of the complicated second derivative terms, so that the contrastive divergence gradient becomes exactly proportional to the gradient of the reconstruction error, which is not the case in general score matching estimator. 
Also, our work  gives  an explicit hierarchical realization of auto-encoder based sampling \cite{alain2014regularized}. The connection with Hopfied network also appears new.

\begin{figure}
	\centering
	\setlength{\fboxrule}{1pt}
	\setlength{\fboxsep}{0cm}
	\subfloat{
		\includegraphics[width=.24\linewidth]{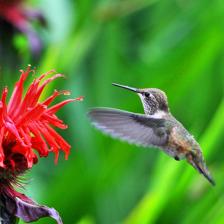}
		\includegraphics[width=.24\linewidth]{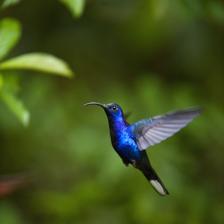}
		\includegraphics[width=.24\linewidth]{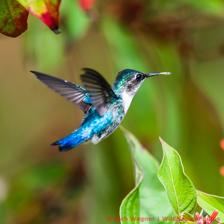}
		\includegraphics[width=.24\linewidth]{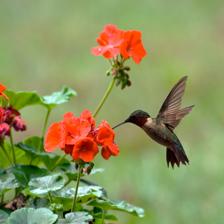}
	}\\[1px]
	\subfloat{
		\includegraphics[width=.24\linewidth]{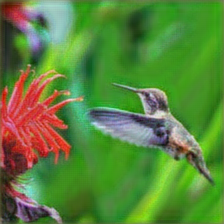}
		\includegraphics[width=.24\linewidth]{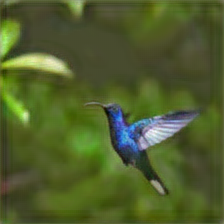}
		\includegraphics[width=.24\linewidth]{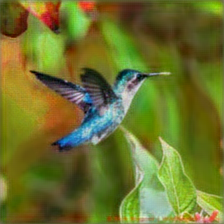}
		\includegraphics[width=.24\linewidth]{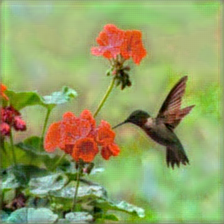}
	}\\[3px]
	\caption{Reconstruction by one-step contrastive divergence. The first row displays 4 of the training images, and the second row displays the corresponding reconstructed images.}
	\label{fig:CD}
\end{figure}

\section{Synthesis and Reconstruction}

We show that the generative ConvNet is capable of learning and generating realistic natural image patterns. Such an empirical proof of concept validates the generative capacity of the model. We also show that contrastive divergence learning can indeed reconstruct the observed images, thus empirically validating Proposition \ref{prop:4}. 

The code in our experiments is based on the MatConvNet package of \cite{matconvnn}. 
%The code and training images can be downloaded from the project page: \url{http://www.stat.ucla.edu/~ywu/GenerativeConvNet/main.html}

Unlike \cite{LuZhuWu2016}, the generative ConvNets in our experiments are learned from scratch without relying on the pre-learned filters of existing ConvNets. 

When learning the generative ConvNet, we grow the layers sequentially. Starting from the first layer, we sequentially add the layers one by one. Each time we learn the model and generate the synthesized images using Algorithm \ref{code:FRAME}.  While learning the new layer of filters,  we  also refine the lower layers of filters by back-propagation.
%we can either fix lower layers of filters while updating the top layer weight and bias parameters according to equations (\ref{eq:generativeGradient}) and (\ref{eq:generativeGradient0}), or we can additionally  refine the lower layer filters by back-propagation at the same time. Both strategies work well for image synthesis. We adopt the latter strategy in our experiments.  

We use $\tilde{M}=16$ parallel chains for Langevin sampling. The number of Langevin iterations between every two consecutive updates of parameters is  $L = 10$.  With each new added layer, the number of learning iterations  is $T = 700$.  We follow the standard procedure to prepare the training images of size $224 \times 224$, whose intensities are within  $[0, 255]$, and we subtract the mean image. We fix $\sigma^2 = 1$ in the reference distribution $q(\I)$. 

For each of the 3 experiments, we use the same set of parameters for all the categories without tuning. 

\subsection{Experiment 1: Generating texture patterns}

We learn a 3-layer generative ConvNet. The first layer has 100 $15 \times 15$ filters with sub-sampling size of 3. The second layer has 64 $5 \times 5$ filters with sub-sampling size of 1. The third layer has 30 $3 \times 3$ filters with sub-sampling size of 1. We learn a generative ConvNet for each category from a single training image. Figure \ref{fig:texture} displays the results. For each category, the first image is the training image, and the rest are 2 of the images  generated by the learning algorithm.

\subsection{Experiment 2:  Generating  object patterns}

 Experiment 1 shows clearly that the generative ConvNet can learn from images without alignment. We can also specialize it to learning aligned object patterns by using a single top-layer filter that covers the whole image. 
 %It is actually a non-stationary FRAME model of the form (\ref{eq:iFRAME}), i.e., a convolutional filter at a fixed position before re-lu non-linearity. 
We learn a 4-layer generative ConvNet from images of aligned objects. The first layer has 100 $7 \times 7$ filters with sub-sampling size of 2. The second layer has 64 $5\times 5$ filters with sub-sampling size of 1. The third layer has 20 $3 \times 3$ filters with sub-sampling size of 1. The fourth layer is a fully connected layer with a single filter that covers the whole image. When growing the layers, we always keep the top-layer single filter, and train it together with the existing layers.   We learn a generative ConvNet for each category, where the number of training images for each category is around 10, and they are collected from the Internet.  Figure \ref{fig:object} shows the results. 
For each category, the first row displays 4 of the training images, and the second row shows 4 of the images generated by the learning algorithm.

\subsection{Experiment 3: Contrastive divergence learns to auto-encode}

We experiment with one-step contrastive divergence learning on a small training set of 10 images collected from the Internet. The ConvNet structure is the same as in experiment 1. For computational efficiency,  we learn all the layers of filters simultaneously.  The number of learning iterations is $T = 1200$. Starting from the observed images, the number of Langevin iterations  is $L = 1$. Figure \ref{fig:CD} shows the results.  The first row displays 4 of the training images, and the second row displays the corresponding auto-encoding reconstructions with the learned parameters. 

\section{Conclusion} 

This paper derives the generative ConvNet from the discriminative ConvNet, and reveals an internal representational structure that is unique among energy-based models. 

%Some of the results in this paper can be mapped to back-propagation in the discriminative ConvNet, but our reinterpretation of them in terms of representation is novel and is richly expansive. Our paper unifies, reconciles or connects the following antagonizing or disparate  pairs:  (1) discriminative ConvNet and generative CongNet, (2) supervised learning and unsupervised learning, (3) exponential family models and  latent variable models, (4) bottom-up filters (operation) and top-down basis functions (representation), (5) synthesis (dream) and reconstruction (memory), (6) hierarchal probability model and hierarchical auto-encoder,  (7) Hopfield attractor network and auto-encoder, (8) Contrastive divergence (learning) and reconstruction (memory). 

The generative ConvNet has the potential to learn from big unlabeled data, either by contrastive divergence or some reconstruction based methods. 

%The generative ConvNet can  be generalized to model dynamic textures and sound data by adding the temporal dimension. 

\section*{Code and data}

The code and training images can be downloaded from the project page: \url{http://www.stat.ucla.edu/~ywu/GenerativeConvNet/main.html}

\newpage

\section*{Acknowledgement}

The code in our work is based on the Matlab code of MatConvNet  of \cite{matconvnn}. We thank the authors for making their code public.

We thank the three reviewers for their insightful comments that have helped us improve the presentation of the paper. We thank Jifeng Dai and Wenze Hu for earlier collaborations related to this work. 

The work is supported by NSF DMS 1310391, ONR MURI N00014-10-1-0933 and DARPA SIMPLEX N66001-15-C-4035.

\bibliography{arXivGConvNet}
\bibliographystyle{icml2016}

\end{document}